\documentclass{tlp}

\newif\ifdraft\drafttrue
\newif\ifinlineref\inlinereffalse
\newif\iffinal\finalfalse
\newif\ifextended\extendedfalse
\newif\ifdotikz\dotikzfalse

\inlinereftrue
\draftfalse
\finaltrue

\def\papertitle{Conflict-driven ASP Solving with External Sources}

\usepackage{times}
\usepackage[scaled]{helvet}
\usepackage{courier}

\usepackage{latexsym}
\usepackage{amsmath}
\usepackage{amssymb}
\usepackage[e]{esvect}
\usepackage{harpoon}
\usepackage{mathdots}

\usepackage{paralist}
\usepackage{wrapfig}
\usepackage{caption}

\usepackage{url}
\makeatletter
\let\UrlSpecialsOld\UrlSpecials
\def\UrlSpecials{\UrlSpecialsOld\do\/{\Url@slash}\do\_{\Url@underscore}}%
\def\Url@slash{\@ifnextchar/{\kern-.11em\mathchar47\kern-.2em}%
    {\kern-.0em\mathchar47\kern-.08em\penalty\UrlBigBreakPenalty}}
\def\Url@underscore{\nfss@text{\leavevmode \kern.06em\vbox{\hrule\@width.3em}}}
\makeatother

\usepackage{color}
\usepackage[vlined,algoruled,rightnl,nofillcomment]{algorithm2e}
\SetAlFnt{\sffamily\scriptsize}

\usepackage{tikz}
\pgfrealjobname{hexcdcl-iclp}
\usetikzlibrary{shapes,arrows,backgrounds,chains,%
matrix,patterns,arrows,decorations.pathmorphing,decorations.pathreplacing,%
positioning,fit,calc,decorations.text,shadows,plotmarks%
}

\iffinal\else
\fi
\ifdraft
\newcommand{\comment}[1]{{\bf\color{blue}{*** #1 ***}}}
\else
\newcommand{\comment}[1]{}
\fi

\usepackage{paralist}
\usepackage{enumerate}
\usepackage{booktabs}
\usepackage{alltt}
\usepackage{caption}

\usepackage{graphicx}
\graphicspath{{figures/}}
\usepackage{subfig}
\usepackage{ifpdf}
\ifpdf
\usepackage{microtype}
\fi

\ifdraft
\usepackage{pdfsync}
\usepackage{srcltx}
\else
\fi

\ifdotikz
\usepackage{tikz}
\pgfrealjobname{hexcdcl}
\usetikzlibrary{shapes,arrows,backgrounds,chains,%
matrix,patterns,arrows,decorations.pathmorphing,decorations.pathreplacing,%
positioning,fit,calc,decorations.text,shadows%
}
\else
\long\def\beginpgfgraphicnamed#1#2\endpgfgraphicnamed{\includegraphics{#1}}
\fi

\newcommand{\qedhere}[0]{\hfill$\Box$}

\newcommand{\leanparagraph}[1]{\smallskip\noindent\textbf{#1}. }

\newenvironment{myitemize}{\begin{list}{$\bullet$}{%
\setlength{\topsep}{0pt}
\setlength{\leftmargin}{0pt}
\setlength{\itemindent}{10pt}}
\parskip=0pt
}
{\end{list}}

\renewcommand{\vec}[1]{{\bf #1}}

\newcommand{\ext}[3]{\ensuremath{\amp{#1}[#2](#3)}}
\DeclareMathOperator{\naf}{not}

\newcommand{\extfun}[1]{\ensuremath{f_{\text{\sl\&}#1}}}

\newcommand{\extsem}[4]{\ensuremath{f_{\text{\sl\&}#1}(#2,#3,#4)}}

\newcommand{\amp}[1]{\ensuremath{\text{\textsl{{\&}}}\!\mathit{#1}}}

\newcommand{\nop}[1]{}

\newcommand\hex{{\sc hex}}
\newcommand\dlvhex{{\sc dlvhex}}

\newcommand{\dlv}[0]{\texttt{DLV}}
\newcommand{\clasp}[0]{{\sc clasp}}
\newcommand{\clingcon}[0]{{\sc clingcon}}
\newcommand{\smodels}[0]{{\sc smodels}}

\newcommand{\T}{\mathbf{T}}
\newcommand{\F}{\mathbf{F}}
\newcommand{\tT}{\mathbf{t}}
\newcommand{\fF}{\mathbf{f}}

\newcommand{\Assignment}{\ensuremath{\mathbf{A}}}
\newcommand{\Program}{\ensuremath{\Pi}}
\newcommand{\ShiftedProgram}{\ensuremath{\mathit{sh}(\Program)}}
\newcommand{\decisionlevel}{\ensuremath{\mathit{dl}}}
\newcommand{\ClarkCompletionNogoods}{\ensuremath{\Delta_{\Program}}}
\newcommand{\SingularLoopNogoods}{\ensuremath{\Theta_{\ShiftedProgram}}}

\newcommand{\ProgramP}{\ensuremath{\hat{\Pi}}}
\newcommand{\ShiftedProgramP}{\ensuremath{\mathit{sh}(\ProgramP)}}
\newcommand{\ClarkCompletionNogoodsP}{\ensuremath{\Delta_{\ProgramP}}}
\newcommand{\SingularLoopNogoodsP}{\ensuremath{\Theta_{\ShiftedProgramP}}}
\newcommand{\LoopNogoodsP}{\ensuremath{\lambda_{\ProgramP}}}

\newcommand{\DynNogoods}{\ensuremath{\nabla}}

\newcommand{\DisjCheckProg}{\ensuremath{\Gamma_{\Program}^\Assignment(C)}}
\newcommand{\sat}{\ensuremath{\mathit{sat}}}
\newcommand{\support}{\ensuremath{\mathit{sup}}}

\newcommand{\Propagation}{\textsf{Propagation}}
\newcommand{\Analysis}{\textsf{Analysis}}
\newcommand{\Select}{\textsf{Select}}
\newcommand{\UnfoundedSet}{\textsf{UnfoundedSet}}
\newcommand{\HEXEval}{\textsf{\hex-Eval}}
\newcommand{\CDNLHEX}{\textsf{\hex-CDNL}}

\newcommand{\CS}{\ensuremath{\mathbf{A}}}
\newcommand{\CSC}{\ensuremath{\mathbf{C}}}

\newcommand{\BA}{\mathit{BA}}

\usepackage{ntheorem}

\newtheorem{theorem}{Theorem}%

\newtheorem{proposition}{Proposition}
\newtheorem{lemma}[theorem]{Lemma}
\newtheorem{definition}{Definition}

{\theorembodyfont{\normalfont}
\newtheorem{example}{Example}
}

{\theorembodyfont{\normalfont}
\theoremheaderfont{\itshape}
\newtheorem*{proofsketch}{Proof (Sketch).}
}

\newcommand{\extproof}[1]{
  \begin{proofsketch}
    #1
  \end{proofsketch}
}

\title[Theory and Practice of Logic Programming]{\papertitle%
}

\author[Eiter et al.]
{Thomas Eiter, Michael Fink, Thomas Krennwallner, and Christoph Redl%
\thanks{This research has been supported by the Austrian Science
    Fund (FWF) project P20840, P20841, and P24090, and by the Vienna
    Science and Technology Fund (WWTF) project ICT 08-020.}\\
  Institut f\"ur Informationssysteme, Technische Universit\"at Wien\\
  \iffinal%
  Favoritenstra\ss{}e\ 9-11, A-1040 Vienna, Austria\\
  \fi
  \email{$\{$eiter,fink,tkren,redl$\}$@kr.tuwien.ac.at}
}

\jdate{}
\pubyear{2012}
\pagerange{\pageref{firstpage}--\pageref{lastpage}}
\doi{}

\begin{document}
\setlength{\abovedisplayskip}{2pt}
\setlength{\abovedisplayshortskip}{2pt}
\setlength{\belowdisplayskip}{2pt}
\setlength{\belowdisplayshortskip}{2pt}

\maketitle

\begin{abstract}
 Answer Set Programming (ASP)
is a well-known problem
solving approach
based on nonmonotonic logic programs and
 efficient solvers. To enable access to external information,
\hex-programs extend programs with \emph{external atoms}, which
 allow for a bidirectional communication between the logic program and
 external sources of computation (e.g., description logic reasoners and
 Web resources).
  Current solvers evaluate \hex-programs by a %
  translation to ASP itself, in which values of external atoms are guessed and
  verified after the ordinary answer set computation.
  This elegant approach does not scale with the number of external
  accesses in general, in particular in presence of nondeterminism
  (which is instrumental for ASP).
  In this paper, we present a novel, native algorithm for evaluating
  \hex-programs which uses learning techniques.
  In particular, we extend conflict-driven ASP solving techniques,
  which prevent the solver from running into the same conflict again,
  from ordinary to \hex-programs.  We show how to gain additional knowledge from external
  source evaluations and how to use
  it in a conflict-driven algorithm.  We first target the uninformed case, i.e., when we have no
  extra information on external sources, and then extend our approach to
  the case where additional meta-information is available.
  Experiments show that learning from external sources can significantly
  decrease both the runtime and the number of considered candidate compatible sets.
\end{abstract}

\section{Introduction}

Answer Set Programming (ASP)
is a
declarative programming approach  \cite{niem-99,mare-trus-99,lifs-2002},
in which solutions to a problem correspond to answer
sets~\cite{gelf-lifs-91} of a logic program, which are computed using an
ASP solver. While this approach has turned out, thanks to expressive and
efficient systems like \smodels{} \cite{simo-niem-02}, \dlv{}
\cite{leon-etal-06-dlv}, ASSAT~\cite{lz2004-aij},
cmodels~\cite{glm2006-jar}, and \clasp{} \cite{gks2012-aij,gkkoss2011-aicom}, to be
fruitful for a range of applications, cf.\ \cite{brew-etal-11-asp},
current trends in distributed systems and the World Wide
Web, for instance, revealed the need for
access to external sources in a program,
ranging from light-weight data access (e.g., XML, RDF, or data bases) to
knowledge-intensive formalisms (e.g., description logics).

To cater for this need, \hex{}-programs~\cite{eist2005} extend ASP
with so called external atoms, through which the user can couple any
external data source with a logic
program.  Roughly, such atoms pass information from the program, given
by predicates and constants, to an external source which
returns output values of an (abstract) function that it computes.
This extension is convenient and has been exploited for
applications in different areas, cf.\ \cite{efiks2011-lpnmr}, and it is also
very expressive since recursive data exchange between the logic program and
external sources is possible. Advanced reasoning
applications like default
reasoning over description logic ontologies \cite{eilst2008-aij,dek2009}
or reasoning over Nonmonotonic Multi-Context
Systems~\cite{be2007,efsw2010-kr} take advantage of it.

Current algorithms for evaluating \hex-programs use a translation
approach and rewrite them to ordinary ASP programs. The idea is to
guess the truth values of external atoms (i.e., whether a
particular fact is in the ``output'' of the external source access) in
a modified program; after computing answer sets, a
compatibility test checks whether the guesses coincide with the actual source behavior.
While elegant, this approach is a bottleneck in advanced applications
including those mentioned above. It does not scale, as
blind guessing leads to an explosion of candidate answer
sets, many of which might fail the compatibility test.
Furthermore, a blackbox view of external sources disables any
pruning of the search space in the ASP translation, and even if
properties would be known, it is sheer impossible
to make use of them in ordinary ASP evaluation  on-the-fly using standard
solvers.

To overcome this bottleneck, a new evaluation method is needed.  In this
paper, we thus present a novel algorithm for evaluating \hex-programs,
described in Section~\ref{sec:learning:algorithms},
which avoids the simple ASP translation approach. It has three key
features.
\begin{myitemize}
\item  First, it natively builds model candidates from first
principles and accesses external sources already during the model
search, which allows to prune candidates early.
\item  Second, it considers
external sources no longer as black boxes, but exploits
meta-knowledge about their internals.
\item And third, it takes up modern SAT and ASP
solving techniques based on \emph{clause learning}~\cite{HandbookOfSAT2009}, which led to very efficient
\emph{conflict-driven} algorithms for answer-set computation~\cite{gks2012-aij,Drescher08conflict-drivendisjunctive},
and extends them to external sources, which is a major contribution of
this work. To this end,
we introduce {\em external behavior learning (EBL)}, which generates conflict clauses (nogoods) after
external source evaluation (Section~\ref{sec:learning:algorithms}).
We do this in Section~\ref{sec:learning}, first in the uninformed case (Section~\ref{sec:learning:uninformed}), where no meta-information
about the external source is available, except that a certain input generates a
certain output. We then exploit meta-information\footnote{Not to be confused with semantically annotated data, which is not considered here.} about external sources
(properties such as monotonicity and functionality) %
to learn even more effective
nogoods which restrict the search space further (Section~\ref{sec:learning:informed}).
\end{myitemize}

We have implemented the new algorithm and incorporated it into
the \dlvhex{} prototype system.\footnote{\url{http://www.kr.tuwien.ac.at/research/systems/dlvhex/}} It is designed in an extensible fashion,
such that the provider of external sources
can specify refined learning functions which exploit specific knowledge about
the source.
Our theoretical work is confirmed by experiments that we
conducted with our prototype on synthetic benchmarks and
programs motivated by real-world applications (Section~\ref{sec:impl}). In several cases,
significant performance improvements compared to the previous algorithm
are obtained, which shows the suitability and potential of the new approach.

\section{Preliminaries}
\label{sec:prelim}

In this section, we introduce syntax and semantics of \hex{}-programs
and,
following~\cite{Drescher08conflict-drivendisjunctive},
conflict-driven SAT and
answer set solving.
We start
with basic definitions.
A (signed) literal is a positive or a negated ground atom
$\T a$ or $\F a$, where ground atom $a$ is of form $p(c_1, \dotsc, c_\ell)$,
with predicate $p$ and function-symbol free ground terms $c_1, \dotsc, c_\ell$,
abbreviated as $p(\vec{c})$.
For a literal $\sigma = \T a$ or $\sigma = \F a$,
let $\overline{\sigma}$ denote its negation,
i.e. $\overline{\T a} = \F a$ and $\overline{\F a} = \T a$.
An assignment $\Assignment$ over a (finite) set of atoms $\mathcal{A}$
is a consistent set of signed literals
$\T a$ or $\F a$, where $\T a$ expresses that~$a \in \mathcal{A}$
is true and
$\F a$ that it is false.

We write $\Assignment^{\T}$ to refer to the set of elements
$\Assignment^{\T} = \{ a \mid \T a \in \Assignment \}$
and $\Assignment^{\F}$ to refer to
$\Assignment^{\F} = \{ a \mid \F a \in \Assignment \}$.
The extension of a %
predicate symbol $q$
wrt. an assignment~$\Assignment$ is
defined as
$\mathit{ext}(q, \Assignment) =
\{ \vec{c} \mid \T q(\vec{c}) \in \Assignment \}$.
Let further~${\Assignment}|_{q}$ be the set of all signed literals over
atoms of form $q(\vec{c})$ in $\Assignment$. For a list $\vec{q} =
q_1,\dotsc,q_k$ of
predicates, we let ${\Assignment}|_{\vec{q}}
= {\Assignment}|_{q_1} \cup \dotsb \cup {\Assignment}|_{q_k}$.

A \emph{nogood} $\{ L_1, \dotsc, L_n \}$ is a set of (signed) literals
$L_i, 1 \le i \le n$.
An assignment $\Assignment$ is a \emph{solution} to a nogood $\delta$
resp.\ a set of nogoods $\Delta$, iff $\delta \not\subseteq \Assignment$
resp.\ $\delta \not \subseteq \Assignment$ for all $\delta \in \Delta$.

\subsection{\hex-Programs}

We briefly recall \hex-programs, which have been introduced
in~\citeN{eist2005} as a generalization of (disjunctive)
extended logic programs under the answer set
semantics~\cite{gelf-lifs-91}; for more details and background, we refer
to~\citeN{eist2005}.

\leanparagraph{Syntax}
\hex-programs extend ordinary ASP programs by \emph{external atoms},
which enable a bidirectional interaction between a program
and external sources of computation.
External atoms have a list of input parameters (constants or predicate names)
and a list of output parameters. Informally,
to evaluate an external atom, the reasoner passes the constants and extensions of the predicates
in the input tuple to the external source
associated with the external atom, which
is plugged into the reasoner. %
The external source
computes %
an output tuple, which %
is matched %
with the output list.
More formally, %
a \emph{ground external atom} is of the form $\ext{g}{\vec{p}}{\vec{c}}$, where
$\vec{p} = p_1, \dotsc, p_k$ are constant input parameters (predicate names or object constants),
and
$\vec{c} = c_1, \dotsc, c_l$ are
constant output terms. %

Ground \hex-programs are then defined similar to ground ordinary ASP programs.

\begin{definition}[Ground \hex-programs]
A ground \hex-program consists of rules of form
\begin{equation*}
  a_1\lor\cdots\lor a_k \leftarrow b_1,\dotsc, b_m, \naf\, b_{m+1},
  \dotsc, \naf\, b_n \ ,
\end{equation*}
where each $a_i$ for $1 \le i \le k$ is a ground atom $p(c_1,\dotsc,c_\ell)$
with constants $c_j$, $1 \le j \le \ell$,
and each~$b_i$ for $1 \le i \le n$ is either a classical ground atom or a ground external atom.%
\footnote{
For simplicity, we do not formally introduce strong negation
but see classical literals of form $\neg a$ as new atoms
together with a constraint which disallows that $a$ and $\neg a$ are simultaneously true.}
\end{definition}

The \emph{head} of a rule $r$ is
$H(r) = \{a_1, \dotsc, a_k \}$ and
the \emph{body}
is $B(r) = \{b_1, \dotsc, b_m,$ $\naf\, b_{m+1}, \dotsc, \naf\, b_n\}$.
We call $b$ or $\naf b$ in a rule body a \emph{default literal};
$B^{+}(r) = \{b_1, \dotsc, b_m\}$ is the \textit{positive body},
$B^{-}(r) = \{b_{m+1}, \dotsc, b_n\}$ is the \textit{negative body}.

In Sections~\ref{sec:learning} and~\ref{sec:impl} we will also make use
of non-ground programs.  However, we restrict our theoretical
investigation to ground programs as suitable safety conditions allow for
application of grounding procedure \cite{eist2006}.
\leanparagraph{Semantics and Evaluation}
The semantics of a ground external atom $\ext{g}{\vec{p}}{\vec{c}}$
wrt. an assignment $\Assignment$ is given by the value of a $1{+}k{+}l$-ary Boolean
\emph{oracle function} $\extfun{g}$ that is defined for all possible values
of $\Assignment$, $\vec{p}$ and $\vec{c}$.  Thus,
$\ext{g}{\vec{p}}{\vec{c}}$ is true relative
to %
$\Assignment$ if and only if it holds that
$\extsem{g}{\Assignment}{\vec{p}}{\vec{c}} = 1$.
Satisfaction of ordinary  rules and ASP programs~\cite{gelf-lifs-91}
is then extended to
\hex-rules and programs in the obvious way, and
the notion of extension $\mathit{ext}(\cdot, \Assignment)$
for external predicates $\amp{g}$ with input lists $\vec{p}$
is naturally defined by
$\mathit{ext}(\amp{g}[\vec{p}], \Assignment) =
\{ \vec{c} \mid \extsem{g}{\Assignment}{\vec{p}}{\vec{c}} = 1\}$.

The answer sets of a \hex-program $\Program$ are
determined by the \dlvhex{} solver using a transformation
to ordinary ASP programs as follows.
Each external atom
$\ext{g}{\vec{p}}{\vec{c}}$
in~$\Program$ is replaced by an ordinary ground \emph{replacement atom}
$e_{\amp{g}[\vec{p}]}(\vec{c})$
and a rule~
$e_{\amp{g}[\vec{p}]}(\vec{c}) \vee \mathit{ne}_{\amp{g}[\vec{p}]}(\vec{c}) \leftarrow$
is added to the program. The answer sets of the resulting \emph{guessing program}~$\ProgramP$
are determined by an ordinary ASP solver and projected
to non-replacement atoms.
However, the resulting
assignments are not necessarily models
of $\Program$, as %
the value of~$\amp{g}[\vec{p}]$ under~$f_{\amp{g}}$
can be different from the one of~$e_{\amp{g}[\vec{p}]}(\vec{c})$.
Each answer set of $\ProgramP$ is thus a \emph{candidate compatible set}
(or \emph{model candidate})
which must be checked against the external sources.
If no discrepancy is found, the model candidate is a
\emph{compatible set} of~$\Program$. More precisely,
\begin{definition}[Compatible Set]
\label{def:compatibleset}
A \emph{compatible set} of a program $\Program$
is an assignment $\Assignment$
\begin{compactenum}[(i)]
\item\label{en:cs1} which is an answer set \cite{gelf-lifs-91} of the \emph{guessing program}
$\ProgramP$, and
\item\label{en:cs2} $\extsem{g}{\Assignment}{\vec{p}}{\vec{c}} = 1$ iff
$\T e_{\amp{g}[\vec{p}]}(\vec{c}) \in \Assignment$ for
all external atoms $\amp{g}[\vec{p}](\vec{c})$ in $\Program$,
i.e. the guessed values coincide with the actual output
under the input from $\Assignment$.
\end{compactenum}
\end{definition}
The compatible sets of $\Pi$ computed by \dlvhex{} include (modulo $A(\Pi)$) all answer
sets of $\Pi$ as defined in \citeN{eist2005} using the
FLP reduct \cite{flp2011-ai}, which we refer to as
FLP-answer sets; with an additional test on candidate answer sets $A$ (which is easily formulated
as compatible set existence for a variant of $\Pi$), the FLP-answer sets can be obtained.  By
default, \dlvhex{} computes compatible sets with smallest true part on
the original atoms; this leads to answer sets as follows.
\begin{definition}[Answer Set]
\label{def:answerset}
An (\dlvhex)  answer set of $\Program$ is any set $S \subseteq \{ \T a \mid a \in A(\Program) \}$
such that \begin{inparaenum}[(i)]
\item $S= \{ \T a \mid a \in A(\Program)\} \cap \Assignment$ for some
compatible set $\Assignment$ of $\Pi$ and
\item $\{ \T a \mid a \in A(\Program)\} \cap \Assignment \not\subset S$ for every
compatible set $\Assignment$ of $\Pi$.
\end{inparaenum}
\end{definition}

The answer sets in Definition~\ref{def:answerset} include all
FLP-answer sets, and in fact often coincide with them (as in all
examples we consider).  Computing the (minimal) compatible
sets is thus a key problem for \hex-programs on which we focus here.

\subsection{Conflict-driven Clause Learning and Nonchronological Backtracking}
Recall that DPLL-style SAT solvers rely on an alternation of drawing deterministic consequences
and guessing the truth value of an atom towards a complete interpretation.
Deterministic consequences are drawn by the basic operation of \emph{unit propagation},
i.e., whenever all but one signed literals of a nogood are satisfied, the last one must be false.
The solver stores an integer \emph{decision level} $\mathit{dl}$, written $@ \mathit{dl}$ as
postfix to the signed literal.
An atom which is set by unit
propagation gets the highest decision level of all already assigned atoms,
whereas guessing increments the current decision level.

Most modern SAT solver are \emph{conflict-driven}, i.e.,
they learn additional nogoods when current assignment violates a nogood.
This prevents the solver from running into the same conflict again.
The learned nogood is determined by initially setting the conflict nogood to the violated one.
As long as it contains multiple literals from the same decision level,
it is resolved with the \emph{reason} of one of these literals, i.e.,
the nogood which implied it.
\begin{example}
\label{ex:cdcl}
Consider the nogoods
\begin{equation*}
\{ \T a, \T b \}, \{ \T a, \T c \}, \{ \F a, \T x, \T y \},
\{ \F a, \T x, \F y \}, \{ \F a, \F x, \T y \},
\{ \F a, \F x, \F y \}
\end{equation*}
and suppose the assignment is $\Assignment = \{ \F a@1, \T b@2, \T c@3,
\T x@4 \}$.  Then the third nogood is unit and implies $\F y@4$,
which violates the fourth nogood $\{ \F a,
\T x, \F y \}$.  As it contains multiple literals ($x$ and $y$) which were
set at decision level $4$, it is resolved with the reason for setting
$y$ to false, which is the nogood $\{ \F a, \T x, \T y \}$. This results
in the nogood $\{ \F a, \T x \}$, which contains the single literal $x$
set at decision level $4$, and thus is the learned nogood.

In standard clause notation, the nogood set corresponds to
$$(\neg a \vee \neg b) \wedge (\neg a \vee \neg c) \wedge (a \vee \neg x \vee \neg y)
\wedge (a \vee \neg x \vee y) \wedge (a \vee x \vee \neg y) \wedge (a \vee x \vee y)$$
and the violated clause is $(a \vee \neg x \vee y)$. It is resolved with $(a \vee \neg x \vee \neg y)$
and results in the learned clause $(a \vee \neg x)$.
\qedhere
\end{example}

State-of-the-art SAT and ASP solvers backtrack then to the second-highest
decision level in the
learned nogood.
In Example~\ref{ex:cdcl}, this is
decision level $1$.
All assignments after decision level $1$ are undone ($\T b@2$,
$\T c@3$, $\T x@4$). Only variable $\F a@1$ remains assigned.
This makes the new nogood $\{ \F a, \T x \}$ unit and derives $\F x$ at
decision level $1$.

\subsection{Conflict-driven ASP Solving}
In this subsection we summarize conflict-driven
(disjunctive) answer-set solving %
\cite{gks2012-aij,Drescher08conflict-drivendisjunctive}.
It corresponds to Algorithm~\ref{alg:hexcdcl} without
Part~\ref{alg:hexcdcl:3}, %
(cf.~Section~\ref{sec:learning:algorithms}, where we also discuss Part~\ref{alg:hexcdcl:3}).
Subsequently, we provide a
summary of the base algorithm; for details we refer
to \citeN{gks2012-aij} and \citeN{Drescher08conflict-drivendisjunctive}.

To employ conflict-driven techniques from SAT solving
in ASP,
 programs are represented as sets of nogoods.
For a program $\Program$, let $A(\Program)$ be the set of all
atoms occurring in $\Program$,
and %
let $\BA(\Program)= \{B(r) \mid r \in \Program\}$ %
be the set of all rule bodies of $\Program$, viewed as fresh atoms.

We first define the set $\gamma(C) = \{ \{\F C\} \cup \{ \tT \ell \mid \ell \in C \} \} \cup
 \{ \{ \T C, \fF \ell \} \mid \ell \in C \}$ of nogoods to encode that a set
$C$ of default literals must be assigned $\T$ or $\F$ in terms of the
conjunction of its elements, where $\tT \naf a = \F a$,
$\tT a = \T a$,
$\fF \naf a = \T a$,
and $\fF a = \F a$.
That is, the conjunction is true iff each literal is true.
Clark's completion $\ClarkCompletionNogoods$ of a program~$\Program$ over
atoms $A(\Program) \cup \BA(\Program)$ is the
set of nogoods
\begin{equation*}
  \ClarkCompletionNogoods = \bigcup\nolimits_{r \in \Program}(\gamma(B(r)) \cup
  \{ \{ \T B(r) \} \cup \{ \F a \mid a \in H(r) \} \}) \enspace .
\end{equation*}
The body of a rule is true iff each literal is true, and
if the body is true, a head literal must also be true.
Unless a program is tight~\cite{Fages94consistencyof}, Clark's
completion does not fully capture the semantics of a program; unfounded
sets may occur, i.e., sets of atoms which only cyclically support each
other, called a \emph{loop}.
Avoidance of unfounded sets requires additional \emph{loop nogoods},
but as there are exponentially many, they are only introduced
on-the-fly.

Disjunctive programs require additional concepts.
Neglecting details, it is common to use
additional nogoods $\SingularLoopNogoods$ derived from
the \emph{shifted program} $\ShiftedProgram$,
which encode the loop formulas of singleton loops;
a comprehensive study is available in~\citeN{Drescher08conflict-drivendisjunctive}.

\nop{
For a disjunctive program $\Program$ the notion of the
\emph{shifted program} $\ShiftedProgram$ is important:

$\ShiftedProgram = \{ a_i \leftarrow B(r), \neg a_1, \dotsc, \neg a_{i - 1},
 \neg a_{i + 1}, \dotsc, \neg a_\ell \mid r \in \Program, H(r) =
 \{ a_1, \dotsc, a_{i - 1}, a_i, a_{i + 1}, \dotsc, a_\ell \} \}$

This leads to the additional nogoods:

$\SingularLoopNogoods = \bigcup_{\overrightarrow{r} \in \ShiftedProgram}
 \gamma(B(\overrightarrow{r})) \cup \{ \delta(a, B(\support_{\ShiftedProgram}(\{a\})))
 \mid a \in \Assignment(\ShiftedProgram) ) \}$

where the \emph{external support} $\support_{\Program}(Y)$ of an atom set $Y$
wrt. a program $\Program$ is defined as the set of rules
$\support_{\Program}(Y) := \{ r \in \Program \mid H(r) \cap Y \not=
 \emptyset, B(r) \cap Y = \emptyset \}$.
}

With these concepts we are ready to describe the basic algorithm for
answer set computation shown in \ref{alg:hexcdcl}.
The algorithm keeps a set
$\ClarkCompletionNogoods \cup \SingularLoopNogoods$ of ``static'' nogoods
(from Clark's completion and from singular loops), and a set $\DynNogoods$ of
``dynamic'' nogoods which are learned from conflicts and unfounded sets
during execution.
While constructing the assignment $\Assignment$, the
algorithm stores for each atom $a \in A(\Program)$ a \emph{decision level}
$\mathit{dl}$. The decision level is initially $0$ and incremented for each
choice. Deterministic consequences of a set of assigned values
have the same decision level as the highest decision level in this set.

The main loop iteratively derives deterministic
consequences using $\Propagation$
trying to complete the assignment.
This includes both unit propagation and unfounded set propagation.
Unit propagation derives $\overline{d}$ if $\delta \setminus \{d\} \subseteq \Assignment$
for some nogood $\delta$,
i.e. all but one literal of a nogood are satisfied, therefore the last one needs to be falsified.
Unfounded set propagation detects atoms which only cyclically support each other
and falsifies them.

Part~\ref{alg:hexcdcl:1} checks if there is a conflict, i.e.
a violated nogood $\delta \subseteq \Assignment$.
If this is the case we need to backtrack.
For this purpose we use $\Analysis$ to compute a learned nogood~$\epsilon$
and a backtrack decision level $k$.
The learned nogood is added to the set of dynamic nogoods, and assignments above
decision level $k$ are undone.
Otherwise, Part~\ref{alg:hexcdcl:2} checks if the assignment is complete.
In this case, a final unfounded set check is necessary due to disjunctive heads.
If the candidate is founded, it is an answer set.
Otherwise we select a violated loop nogood~$\delta$
from the set $\LoopNogoodsP(U)$ of all loop nogoods for an unfounded set $U$
(for the definition see~\citeNP{Drescher08conflict-drivendisjunctive}),
we do conflict analysis and backtrack.
If no more deterministic consequences can be derived and the
assignment is still incomplete, we need to guess in Part~\ref{alg:hexcdcl:4}
and increment the decision level.
The function $\Select$ implements a variable selection heuristic.
In the simplest case it chooses an arbitrary yet unassigned variable,
but state-of-the-art heuristics are more sophisticated.
E.g., \citeN{Goldberg2007} prefer variables
which are involved in recent conflicts.

\nop{
The \emph{independent satisfaction} of a rule $r$ wrt. to a set of
literals $Y$ is defined as the set:
$\sat_r(Y) := \{ \F B(r) \} \cup \{ \T a \mid a \in H(r) \setminus Y \}$;
i.e., the set of literals such that the truth of any of them
immediately satisfies $r$ independently from $Y$.

Moreover, for components $C$ of disjunctive
programs $\Program$ the following
set of nogoods is important for the minimality check:

\begin{eqnarray*}
\DisjCheckProg &=& \{
   \{ \T a \mid a \in H(r) \cap \Assignment^{\T}\} \cup
   \{ \F a \mid a \in B(r) \cap C\} \mid r \in \Program, \sat_r(C) \cap \Assignment = \emptyset \} \cup \\
		&&	\{ \{ \F a \mid a \in C \cap \Assignment^{\T} \} \}
\end{eqnarray*}
}

\section{Algorithms for Conflict-driven \hex{}-Program Solving}
\label{sec:learning:algorithms}

We present now our new, genuine algorithms
for \hex-program evaluation.
They are based on~\citeN{Drescher08conflict-drivendisjunctive},
but integrate additional novel learning techniques
to capture the semantics of external atoms.
The term \emph{learning} refers to the process of adding further
nogoods to the nogood set as the search space is explored.
They are classically derived from conflict situations to
avoid similar conflicts during further search, as
described above.

We add a second type of learning which captures the behavior of external
sources, called \emph{external behavior learning} (EBL).
Whenever an external atom is evaluated, the algorithm might learn
from the call.
If we have no further information about the internals of a source,
we may learn only very general input-output-relationships, if we have more information
we can learn more effective nogoods.
In general, we can associate a \emph{learning-function} with each external source.
For the sake of introducing the evaluation algorithms, however, in this section
we abstractly consider a set of nogoods learned from the evaluation of some external
predicate with input list $\amp{g}[\vec{p}]$, if evaluated under an assignment
$\Assignment$,
denoted by $\Lambda(\amp{g}[\vec{p}], \Assignment)$.
The next section will provide definitions %
of particular nogoods that can be learned for various types of external sources, i.e.,
to instantiate $\Lambda(\cdot,\cdot)$.
The crucial requirement for learned nogoods is \emph{correctness},
which intuitively holds if the nogood can be added without eliminating compatible sets.
\begin{definition}[Correct Nogoods]
A nogood $\delta$ is \emph{correct wrt. a program $\Program$}, if all compatible
sets of $\Program$ are solutions to $\delta$.
\end{definition}
\setlength{\algomargin}{0mm}
\SetAlCapSkip{0ex}
\SetAlCapHSkip{0ex}
\SetVlineSkip{1mm}
\SetAlgoSkip{}
\SetInd{0.5mm}{3.25mm}
\begin{figure}[t]
\begin{minipage}[t]{0.47\textwidth}
\vspace{0pt} %
\begin{algorithm}[H]
  \SetAlgoRefName{\HEXEval}
  \SetAlgoCaptionSeparator{}
  \caption{}
\label{alg:hexeval}
  \DontPrintSemicolon
  \SetAlgoVlined

\KwIn{A \hex{}-program $\Program$}
\KwOut{All answer sets of $\Program$}
 $\ProgramP \leftarrow \Program$ with ext. atoms $\ext{g}{\vec{p}}{\vec{c}}$
 replaced by $e_{\amp{g}[\vec{p}]}(\vec{c})$\;
 Add guessing rules for all replacement atoms to $\ProgramP$\;
$\DynNogoods \leftarrow \emptyset$ \tcp*[f]{set of dynamic nogoods}\;
$\Gamma \leftarrow \emptyset$ \tcp*[f]{set of all compatible sets}\;
\nlset{(a)}{%
\label{alg:hexeval:0a}%
\While{$\mathbf{C} \not= \bot$}{%
  \Indm
	$\mathbf{C} \leftarrow \bot$\;
	$\mathit{inconsistent} \leftarrow \mathit{false}$\;
	\nlset{(b)}{%
	\label{alg:hexeval:0b}%
	\While{$\mathbf{C} = \bot$ and $\mathit{inconsistent} = \mathit{false}$}{%
		\nlset{(c)}{\label{alg:hexeval:1}%
                  $\CS \leftarrow$$\CDNLHEX($\Program$, $\ProgramP,$ \DynNogoods$)\;}
		\lIf{$\CS = \bot$}{%
			$\mathit{inconsistent} \leftarrow \mathit{true}$\;
		}\Else{
			$\mathit{compatible} \leftarrow \mathit{true}$\;
			\nlset{(d)}{\label{alg:hexeval:2}
			\For{all external atoms $\amp{g}[\vec{p}]$ in $\Program$}{
				Evaluate $\amp{g}[\vec{p}]$ under~$\CS$\;
				\nlset{(e)}{\label{alg:hexeval:3}$\DynNogoods \leftarrow \DynNogoods \cup \Lambda(\amp{g}[\vec{p}], \CS)$\;}

				Let $\CS^{\amp{g}[\vec{p}](\vec{c})} \,{=}\, 1 \Leftrightarrow \T e_{\amp{g}[\vec{p}](\vec{c})} \in \CS$\;
				\If{$\exists \vec{c}\colon\extsem{g}{\CS}{\vec{p}}{\vec{c}} \neq \CS^{\amp{g}[\vec{p}](\vec{c})}$}{
					Add $\CS$ to $\DynNogoods$\;
                                        $\mathit{compatible} \leftarrow \mathit{false}$\;
				}
			}
			}

			\lIf{$\mathit{compatible}$}{
				$\mathbf{C} \leftarrow \CS$\;
			}
		}
	}}
	\If{$\mathit{inconsistent} = \mathit{false}$}{
		\tcp*[f]{$\mathbf{C}$ is a compatible set of $\Program$}\;
		$\DynNogoods \leftarrow \DynNogoods \cup \{ \mathbf{C} \}$ and $\Gamma \leftarrow \Gamma \cup \{ \mathbf{C} \}$\;
	}
}}
\Return $\subseteq$-minimal $\left\{ \{ \T a \in \CS \mid a \in A(\Program) \} \mid \CS \in \Gamma \right\}$\;
\end{algorithm}
\end{minipage}
\hfill
\begin{minipage}[t]{0.51\textwidth}
\vspace{0pt} %
\begin{algorithm}[H]
  \SetAlgoRefName{\CDNLHEX}
  \SetAlgoCaptionSeparator{}
  \caption{}
\label{alg:hexcdcl}
  \DontPrintSemicolon
  \SetAlgoVlined

\KwIn{A program $\Program$, its guessing program $\ProgramP$,
a set of correct nogoods $\DynNogoods\!$ of $\!\Pi$}
\KwOut{An answer set of $\ProgramP$ (candidate for a compatible set of~$\Program$) which is a solution to all nogoods $d \in \DynNogoods$, or $\bot$ if none exists}

$\Assignment \leftarrow \emptyset$ \hspace*{-0.5ex} \tcp*[f]{$\!$over
  $A(\ProgramP)\,{\cup}\, \BA(\ProgramP)\,{\cup}\,
  \BA(\ShiftedProgramP)\!$} \;
$\decisionlevel \leftarrow 0$ \tcp*[f]{decision level} \;
\While{true}{%
  \Indm
  $(\Assignment, \DynNogoods) \leftarrow \Propagation(\ProgramP, \DynNogoods, \Assignment)$\;
  \nlset{(a)}{\label{alg:hexcdcl:1}}%
  \uIf{$\delta \subseteq \Assignment$ for some $\delta \in \ClarkCompletionNogoodsP \cup \SingularLoopNogoodsP \cup \DynNogoods$}{
    \lIf{$\decisionlevel = 0$}{%
      \Return{$\bot$}\;%
    }
    $(\epsilon, k) \leftarrow \Analysis(\delta, \ProgramP, \DynNogoods, \Assignment)$\;
    $\DynNogoods \leftarrow \DynNogoods \cup \{ \epsilon \}$ and $\decisionlevel \leftarrow k$\;
    $\Assignment \leftarrow \Assignment \setminus \{ \sigma \in \Assignment \mid k < \decisionlevel(\sigma) \}$\;
  }%
  \nlset{(b)}{\label{alg:hexcdcl:2}}%
  \uElseIf{$\Assignment^{\T} {\cup} \Assignment^{\F} {=} A(\ProgramP) {\cup} \BA(\ProgramP) {\cup} \BA(\ShiftedProgramP)$}{
    $U \leftarrow \UnfoundedSet(\ProgramP, \Assignment)$\;
    \uIf{$U \not= \emptyset$}{%
      let $\delta \in \LoopNogoodsP(U)$ such that $\delta \subseteq \Assignment$\;
      \lIf{$\{\sigma \in \delta \mid 0 < \decisionlevel(\sigma)\} = \emptyset$}{%
        \Return{$\bot$}\;%
      }%
      $(\epsilon, k) \leftarrow \Analysis(\delta, \ProgramP, \DynNogoods, \Assignment)$\;
      $\DynNogoods \leftarrow \DynNogoods \cup \{\epsilon\}$ and $\decisionlevel \leftarrow k$\;
      $\Assignment \leftarrow \Assignment \setminus \{\sigma \in \Assignment \mid k < \decisionlevel(\sigma)\}$\;
    }\lElse{
      \Return{$\Assignment^{\T} \cap A(\ProgramP)$}\;
    }
  }
  \nlset{(c)}{\label{alg:hexcdcl:3}}%
  \uElseIf{Heuristic decides to evaluate $\amp{g}[\vec{p}]$}{%
    Evaluate $\amp{g}[\vec{p}]$  under~$\CS$ and set
    $\DynNogoods \leftarrow \DynNogoods \cup \Lambda(\amp{g}[\vec{p}], \Assignment)$\;
  }%
  \nlset{(d)}{\label{alg:hexcdcl:4}}%
  \Else{%
    $\sigma \leftarrow \Select(\ProgramP, \DynNogoods,\Assignment)$ and $\decisionlevel \leftarrow \decisionlevel + 1$\;
    $\Assignment \leftarrow \Assignment \circ (\sigma)$\;
  }%
}
\end{algorithm}
\end{minipage}
\end{figure}

In our subsequent exposition we assume that the program $\Program$ is clear from
the context.
The overall approach consists of two parts.  First,
\ref{alg:hexcdcl}  computes model candidates;
it is essentially an ordinary ASP solver, but includes calls to external
sources in order to learn additional nogoods. The external calls
in this algorithm are not required for correctness of the algorithm, but may influence
performance dramatically as discussed in Section~\ref{sec:impl}.
Second, Algorithm~\ref{alg:hexeval} uses Algorithm~\ref{alg:hexcdcl}
to produce model candidates and checks each of them against
the external sources (followed by a minimality check).
Here, the external calls are crucial for correctness of the algorithm.

For computing a model candidate, \ref{alg:hexcdcl} basically
employs the conflict-driven approach presented in \citeN{Drescher08conflict-drivendisjunctive}
as summarized in Section~\ref{sec:prelim},
where the main difference is the addition of Part~\ref{alg:hexcdcl:3}.
Our extension is driven by the following idea: whenever (unit and unfounded set)
propagation does not derive any further atoms and the assignment is
still incomplete, the algorithm possibly evaluates external atoms
(driven by a heuristic) instead of simply guessing truth values.
This might lead to the addition of new nogoods, which can in turn
cause the propagation procedure to derive further atoms.
Guessing of truth values only becomes necessary if no deterministic
conclusions can be drawn and the evaluation of external atoms does not yield
further nogoods; guessing also occurs if
the heuristic does not decide to evaluate.

For a more formal treatment,
let $\mathcal{E}$ be the set of all external predicates with input list
that occur in $\Program$, and let $\mathcal{D}$ be the set of all signed
literals over atoms in %
$A(\Program)\cup A(\ProgramP)\cup \BA(\ProgramP)$.
Then, a \emph{learning function} for $\Program$ is a mapping
$\Lambda: \mathcal{E} \times 2^{\mathcal{D}} \mapsto 2^{2^{\mathcal{D}}}$.
We extend our notion of correct nogoods to correct learning
functions $\Lambda(\cdot, \cdot)$, as follows:
\begin{definition}
A learning function $\Lambda$
is \emph{correct} for a program $\Program$, iff
all~$d \in \Lambda(\amp{g}[\vec{p}], \Assignment)$
are correct for $\Program$, for all
$\amp{g}[\vec{p}]$
in $\mathcal{E}$ and
$\Assignment \in 2^{\mathcal{D}}$.
\end{definition}
Restricting to
learning functions %
that are correct for $\Program$, the following results hold.
\begin{proposition}
\label{prop:hexcdcl}
If for input $\Pi$, $\ProgramP$ and $\DynNogoods$,
\ref{alg:hexcdcl} returns
 \begin{inparaenum}[(i)]
 \item
 \label{prop:hexcdcl:correctness}
an interpretation $\Assignment$, then $\Assignment$ is an answer set of
$\ProgramP$ and a solution to $\DynNogoods$;
 \item
 \label{prop:hexcdcl:completeness}
$\bot$, then $\Program$ has no compatible set that is a
solution to $\DynNogoods$.
\end{inparaenum}
\end{proposition}%
\extproof{
(\ref{prop:hexcdcl:correctness})
The proof mainly follows \cite{Drescher08conflict-drivendisjunctive}.
In our algorithm we have potentially more nogoods, which can never produce
further answer sets but only eliminate them. Hence, each produced
interpretation $\Assignment$ is an answer set of $\ProgramP$.
(\ref{prop:hexcdcl:completeness})
By completeness of \citeN{Drescher08conflict-drivendisjunctive}
we only need to justify that adding
$\Lambda(\amp{g}[\vec{p}], \Assignment)$
after evaluation of $\amp{g}[\vec{p}]$
does not eliminate compatible sets of $\Program$.
For this purpose we need to show that when one of the added nogoods
fires,
the interpretation is incompatible with the external sources anyway.
But this follows from the correctness of $\Lambda(\cdot, \cdot)$
and (for derived nogoods) from the completeness of \citeN{Drescher08conflict-drivendisjunctive}.
\hfill $\Box$
}

The basic idea of \ref{alg:hexeval} is to compute
all compatible sets of $\Program$ by the loop at \ref{alg:hexeval:0a}
and checking subset-minimality  afterwards.
For computing compatible sets, the loop at \ref{alg:hexeval:0b}
uses \ref{alg:hexcdcl} to compute answer sets of
$\ProgramP$ in \ref{alg:hexeval:1}, i.e.,
candidate compatible sets of~$\Program$, and subsequently
checks compatibility for each external atom in \ref{alg:hexeval:2}.  Here
the external calls are crucial for correctness.
However, different from the translation approach, the
external source evaluation serves not only for compatibility checking, but
also for generating additional dynamic nogoods~$\Lambda(\amp{g}[\vec{p}],
\CS)$ in Part~\ref{alg:hexeval:3}.
We have the following result.
\begin{proposition}
\label{prop:hexeval}
\ref{alg:hexeval} computes all answer sets of $\Program$.
\end{proposition}
\extproof{
We first show that the loop at \ref{alg:hexeval:0b} yields after
termination a compatible set~$\CSC$ of $\Program$  that is a solution of
$\DynNogoods$
at the stage of entering the
loop iff such a compatible set does exist, and yields $\CSC=\bot$  iff no
such compatible set exists.

Suppose that %
$\CSC \not= \bot$ after the loop. Then $\CSC$ was assigned $\Assignment
\neq \bot$, which was returned by \CDNLHEX($\Program$,
$\ProgramP$, $\DynNogoods$). From Proposition \ref{prop:hexcdcl} (\ref{prop:hexcdcl:completeness})
it follows that $\CSC$ is an answer set of $\ProgramP$ and a solution
to $\DynNogoods$. Thus \eqref{en:cs1} of
Definition~\ref{def:compatibleset} holds.
As $\mathit{compatible} = \mathit{true}$, the
for loop guarantees the compatibility with the external sources in \eqref{en:cs2} of
Definition~\ref{def:compatibleset}:
if some source output on input from $\CSC$ is not compatible with the guess,
$\CSC$ is rejected (and added as nogood).
Otherwise $\CSC$ coincides with the behavior of the
external sources, i.e., it satisfies $(ii)$ of Definition~\ref{def:compatibleset}.
Thus, $\CSC$ is a compatible set of $\Program$ wrt.\, $\DynNogoods$
at call time. As only correct nogoods are added to $\DynNogoods$, it is
also a compatible set of $\Program$ wrt. the initial set $\DynNogoods$.

Otherwise, after the loop $\CSC=\bot$. Then $\mathit{inconsistent}
=\mathit{true}$, which means that the call \CDNLHEX($\Program$,
$\ProgramP$, $\DynNogoods$) returned $\bot$.
By Proposition \ref{prop:hexcdcl} (\ref{prop:hexcdcl:completeness})
there is no answer set of $\ProgramP$ which is a solution to $\DynNogoods$.
As only correct nogoods were added to $\DynNogoods$, there exists also
no answer set of $\ProgramP$ which is a solution to the original set $\DynNogoods$.
Thus the loop at \ref{alg:hexeval:0b}  operates as desired.

The loop
at \ref{alg:hexeval:0a} then enumerates one by one all compatible sets and terminates:
the update of $\DynNogoods$ with $\CSC$ prevents recomputing
$\CSC$, and thus the number of compatible sets decreases.
As by Definition~\ref{def:answerset} the answer sets of $\Pi$ are the
compatible sets with subset-minimal true
part of original literals, the overall algorithm correctly outputs all answer
sets of $\Program$.
\hfill $\Box$
}

\begin{example}\label{ex:algorithmexample}
  Let $\amp{\mathit{empty}}$ be an external atom with one (nonmonotonic)
  predicate input~$p$, such that its output is $c_0$ if the extension
  of~$p$ is empty and $c_1$ otherwise.
  Consider the program~$\Program_e$ consisting of the rules
  \begin{align*}
    p(c_0).\ \ dom(c_0).\ \ dom(c_1).\ \ dom(c_2). \quad
    p(X) \leftarrow \mathit{dom}(X), \ext{\mathit{empty}}{p}{X}
  \end{align*}
  Algorithm~\ref{alg:hexeval} transforms $\Program_e$ into the guessing program $\hat{\Program}_e$:
  \begin{align*}
    p(c_0).\ \ dom(c_0).\ \ dom(c_1).\ \ dom(c_2). \quad
    p(X) & \leftarrow \mathit{dom}(X), e_{\amp{\mathit{empty}}[p]}(X). \\
    e_{\amp{\mathit{empty}}[p]}(X) \vee
    \mathit{ne}_{\amp{\mathit{empty}}[p]}(X) & \leftarrow \mathit{dom}(X).
  \end{align*}
  The traditional evaluation strategy without learning will then produce
  $2^3$ model candidates in \ref{alg:hexcdcl}, which are
  subsequently checked in \ref{alg:hexeval}.  For instance,
  the guess $\left\{ \T \mathit{ne}_{\amp{\mathit{empty}}[p]}(c_0), \T
    e_{\amp{\mathit{empty}}[p]}(c_1), \T
    \mathit{ne}_{\amp{\mathit{empty}}[p]}(c_2) \right\}$ leads to the model candidate
  $\left\{ \T \mathit{ne}_{\amp{\mathit{empty}}[p]}(c_0), \T
    e_{\amp{\mathit{empty}}[p]}(c_1), \T
    \mathit{ne}_{\amp{\mathit{empty}}[p]}(c_2), \T p(c_1) \right\}$ (neglecting
  false atoms and facts).  This is also the only model candiate which
  passes the compatibility check:~$p(c_0)$ is always true, and
  therefore $e_{\amp{\mathit{empty}}[p]}(c_1)$ must also be true due to
  definition of the external atom. This allows for deriving $p(c_1)$ by
  the first rule of the program. All other atoms are false due to
  minimality of answer sets. \qedhere
\end{example}

The effects of the additionally learned nogoods will be discussed in
Section~\ref{sec:learning} after having formally specified concrete
$\Lambda(\amp{g}[\vec{p}], \Assignment)$ for various types of external
sources.

\section{Nogoods for External Behavior Learning}
\label{sec:learning}

We  now discuss nogoods generated for external behavior
learning (EBL) in detail.
EBL is triggered by external source evaluations
instead of conflicts.
The basic idea is to integrate knowledge about the
external source behavior into the program to guide the search.
The program evaluation then starts with an empty set
of learned nogoods and the preprocessor generates a
guessing rule for each ground external atom, as discussed
in Section~\ref{sec:prelim}. Further nogoods
are added during the evaluation as more information about
external sources becomes available.
This is in contrast to
traditional
evaluation, where
external atoms are assigned arbitrary truth values
which are checked only after the assignment was completed.

We will first show how to construct useful learned nogoods
after evaluating external atoms,
if we have no further information about the internals of external sources,
called \emph{uninformed learning}.
In this case we can only learn simple input/output relationships.
Subsequently we consider \emph{informed learning}, where additional
information about properties of external sources is available. This
allows for using more elaborated learning strategies.

\subsection{Uninformed Learning}
\label{sec:learning:uninformed}
We first assume that we do not have
information about the internals and consider external sources as black
boxes.
Hence, we can just apply very general rules for learning:
whenever an external predicate with
input list $\amp{g}[\vec{p}]$ is
evaluated under an assignment $\Assignment$, we learn that
the input $\Assignment|_{\vec{p}}$ for $\vec{p}=p_1,\dotsc,p_n$ to the
external atom $\amp{g}$ produces the output $\mathit{ext}(\amp{g}[\vec{p}], \Assignment)$.
This can be formalized as the following set of nogoods.
\begin{definition}
\label{def:extlearn:general}
The learning function for a general external predicate with input list
$\amp{g}[\vec{p}]$ in program $\Program$ under assignment $\Assignment$ is defined as
\begin{equation*}\label{eqn:general}
\Lambda_g(\amp{g}[\vec{p}], \Assignment) =
		\left\{  \Assignment|_{\vec{p}} \cup
		\{ \F e_{\amp{g}[\vec{p}]}(\vec{c}) \}
                \mid \vec{c} \in \mathit{ext}(\amp{g}[\vec{p}],
                \Assignment)\right\} \enspace .
\end{equation*}
\end{definition}

In the simplest case, an external atom has no input and
the learned nogoods are unary, i.e., of the form
$\{ \F e_{\amp{g}[]}(\vec{c}) \}$.
Thus, it is learned that certain tuples are in the output of the
external source, i.e. they must not be false.
For external sources with input predicates, the added rules encode the
relationship between the output tuples and the provided input.
\begin{example}[ctd.]\label{ex:learningnogoods}
  Recall $\Program_e$ from Example~\ref{ex:algorithmexample}.  Without
  learning, the algorithms produce $2^3$ model candidates and check them
  subsequently. It turns out that
  EBL allows for falsification
  of some of the guesses without actually evaluating the external atoms.
  Suppose the reasoner first tries  the guesses containing literal $\T
  e_{\amp{\mathit{empty}}[p]}(c_0)$.  While they are checked against the
  external sources, the described learning function allows for adding the
  externally learned nogoods shown in Table~\ref{tab:learnednogoods}.
        \begin{table}[t]
          \centering
          \small
          \caption{Learned Nogoods of Example~\ref{ex:learningnogoods}}\label{tab:learnednogoods}
          \begin{tabular}{ll}
            \toprule\toprule
            \normalsize Guess & \normalsize  Learned Nogood \\
            \midrule
            $\left\{
            \begin{array}{l}
              \T e_{\amp{\mathit{empty}}[p]}(c_0), \T
              \mathit{ne}_{\amp{\mathit{empty}}[p]}(c_1), \\
              \T \mathit{ne}_{\amp{\mathit{empty}}[p]}(c_2)
            \end{array}
            \right\}$
            & $\{ \T p(c_0), \F p(c_1), \F p(c_2), \F e_{\amp{\mathit{empty}}[p]}(c_1) \}$ \\[3mm]
            $\left\{
            \begin{array}{l}
              \T e_{\amp{\mathit{empty}}[p]}(c_0), \T
              \mathit{ne}_{\amp{\mathit{empty}}[p]}(c_1),\\
              \T e_{\amp{\mathit{empty}}[p]}(c_2), p(c_2)
            \end{array}
            \right\}$
            & $\{ \T p(c_0), \F p(c_1), \T p(c_2), \F e_{\amp{\mathit{empty}}[p]}(c_1) \}$ \\[3mm]
            $\left\{
              \begin{array}{l}
                \T e_{\amp{\mathit{empty}}[p]}(c_0), \T
                e_{\amp{\mathit{empty}}[p]}(c_1),\\
                \T \mathit{ne}_{\amp{\mathit{empty}}[p]}(c_2), p(c_1)
              \end{array}
            \right\}$
            & $\{ \T p(c_0), \T p(c_1), \F p(c_2), \F e_{\amp{\mathit{empty}}[p]}(c_1) \}$ \\[3mm]
            $\left\{
              \begin{array}{l}
                \T e_{\amp{\mathit{empty}}[p]}(c_0), \T
                e_{\amp{\mathit{empty}}[p]}(c_1),\\
                \T e_{\amp{\mathit{empty}}[p]}(c_2), p(c_1), p(c_2)
              \end{array}
            \right\}$
            & $\{ \T p(c_0), \T p(c_1), \T p(c_2), \F e_{\amp{\mathit{empty}}[p]}(c_1) \}$ \\
            \bottomrule\bottomrule
          \end{tabular}
        \end{table}
	Observe that the combination $\T p(c_0), \F p(c_1), \F p(c_2)$
        will be reconstructed also for different
        choices of the guessing variables. As $p(c_0)$ is a fact, it is
        true independent of the choice between
        $e_{\amp{\mathit{empty}}[p]}(c_0)$
        and~$\mathit{ne}_{\amp{\mathit{empty}}[p]}(c_0)$.  E.g., the
        guess $\F e_{\amp{\mathit{empty}}[p]}(c_0),$ $\F
        e_{\amp{\mathit{empty}}[p]}(c_1),$ $\F
        e_{\amp{\mathit{empty}}[p]}(c_2)$ leads to the same extension of
        $p$.  This allows for reusing the nogood, which is immediately
        invalidated without evaluating the external atoms.
	Different guesses with the same input to an external source
        allow for reusing  learned nogoods, at the latest when the
        candidate is complete, but before the external source is
        called for validation.  However, very often learning allows for
        discarding guesses even earlier.  For instance, we can derive $\{ \T p(c_0), \F
        e_{\amp{\mathit{empty}}[p]}(c_1) \}$ from the nogoods
        above   in 3 resolution
        steps. Such derived nogoods will be learned after running into a
        couple of conflicts.  %
        We can derive $\T
        e_{\amp{\mathit{empty}}[p]}(c_1)$ from
        $p(c_0)$ even before the truth value of $\F e_{\amp{\mathit{empty}}[p]}(c_1)$ is set,
        i.e., external learning guides the search while the traditional
        evaluation algorithm considers the behavior of external sources
        only during postprocessing. \qedhere
\end{example}

For the next result, let $\Program$ be a program which contains an
external atom of form $\amp{g}[\vec{p}](\cdot)$. %
\begin{lemma}
\label{lem:correctness:uninformed}
For all assignments $\Assignment$, the nogoods
$\Lambda_g(\amp{g}[\vec{p}], \Assignment)$
(Def.~\ref{def:extlearn:general}) are correct wrt. $\Program$.
\end{lemma}
\extproof{
The added nogood for
an output tuple $\vec{c} \in \mathit{ext}(\amp{g}[\vec{p}], \Assignment)$
contains $\Assignment|_{\vec{p}}$
and the negated replacement atom $\F e_{\amp{g}[\vec{p}]}(\vec{c})$.
If the nogood fires, then the guess was wrong
as the replacement atom is guessed false
but the tuple $(\vec{c})$ is in the output.
Hence, the interpretation is not compatible and
cannot be an answer set anyway.
\qedhere
}

\subsection{Informed Learning}
\label{sec:learning:informed}
The learned nogoods of the above form can become quite large as
they include the whole input to the
external source.
However, known properties of external sources can be
exploited in order to learn smaller and more general nogoods.
For example, if one of the input parameters of an external source
is monotonic, it is not necessary to include information about false atoms
in its extension,
as the output will not shrink given larger input.

Properties for informed learning can be stated on the level
of either \emph{predicates} or individual \emph{external atoms}.
The former means that all usages of the predicate
have the property. To understand this, consider predicate
$\amp{\mathit{union}}$ which takes two predicate inputs~$p$ and $q$
and computes the set of all elements which are in at least one
of the extensions of~$p$ or~$q$. It will be \emph{always} monotonic in
both parameters, independently of
its usage in a program.
While an external source may lack a property
in general, it may hold for particular usages.
\begin{example} %
Consider an external atom
$\ext{\mathit{db}}{r_1,\dotsc,r_n,\mathit{query}}{\vec{X}}$
as an interface to an SQL query processor, which
evaluates a given query (given as string) over tables (relations) provided by predicates $r_1,\dotsc,r_n$.
In general, the atom will be nonmonotonic, but for special
queries (e.g., simple selection of all tuples), it will be monotonic.
\qedhere
\end{example}

Next, we discuss two particular cases of informed learning which
customize the default learning function for generic external sources by
exploiting properties of external sources, and finally present examples
where the learning of user-defined nogoods might be useful.

\leanparagraph{Monotonic Atoms}%
A parameter $p_i$ of an external atom $\amp{g}$ is called \emph{monotonic}, if
$\extsem{g}{\Assignment}{\vec{p}}{\vec{c}} = 1$ implies
$\extsem{g}{\Assignment'}{\vec{p}}{\vec{c}} = 1$
for all~$\Assignment'$ with $\Assignment'|_{p_i} \supseteq \Assignment|_{p_i}$
and $\Assignment'|_{p'} = \Assignment|_{p'}$ for all other $p' \not= p_i$.
The learned nogoods
$\Lambda(\amp{g}[\vec{p}], \Assignment)$
after evaluating $\amp{g}[\vec{p}]$
are not required to include $\F p_i(t_1, \dotsc, t_\ell)$
for monotonic $p_i \in \vec{p}$.
That is, for an external predicate with input list $\amp{g}[\vec{p}]$
with monotonic input parameters $\vec{p_m} \subseteq \vec{p}$
and nonmonotonic parameters $\vec{p_n} = \vec{p} \setminus \vec{p_m}$,
the set of learned nogoods can be restricted as follows.
\begin{definition}
\label{def:extlearn:monotonic}
The learning function for an external predicate $\amp{g}$ with input list
$\vec{p}$ in program~$\Program$ under assignment $\Assignment$,
such that $\amp{g}$ is monotonic in $\vec{p_m} \subseteq \vec{p}$, is defined as
\begin{equation*}\label{eqn:monotonic}
\Lambda_m(\amp{g}[\vec{p}], \Assignment) =
		\left\{%
		\{ \T a \in \Assignment|_{\vec{p_m}} \} \cup
		\Assignment|_{\vec{p_n}}
		\cup \{ \F e_{\amp{g}[\vec{p}]}(\vec{c}) \}
		\vphantom{\bigcup\nolimits_{p_i \not\in M}}
		\mid \vec{c} \in \mathit{ext}(\amp{g}[\vec{p}],
                \Assignment) \right\} \enspace .
\end{equation*}
\end{definition}
\begin{example}\label{ex:learningmonotonicity}
  Consider the external atom $\ext{\mathit{diff}}{p,q}{X}$ which
  computes the set of all elements $X$ that are in the extension of $p$,
  but not in the extension of $q$.  Suppose it is evaluated under
  $\Assignment$, s.t. $\mathit{ext}(p, \Assignment) = \{\T p(a), \T p(b),
  \F p(c)\}$ and $\mathit{ext}(q, \Assignment) = \{\F q(a), \T q(b), \F
  q(c)\}$.  Then the output of the atom is
  $\mathit{ext}(\amp{\mathit{diff}}[p, q], \Assignment) = \{ a \}$
  and the (only) naively learned nogood is $\{ \T p(a), \T p(b), \F
  p(c), \F q(a), \T q(b), \F q(c), \F e_{\amp{\mathit{diff}[p, q]}}(a)
  \}$.
  However, due to monotonicity of $\amp{\mathit{diff}}[p, q]$ in $p$, it
  is not necessary to include $\F p(c)$ in the nogood; the output of the
  external source will not shrink even if $p(c)$ becomes true.
  Therefore the (more general) nogood $\{ \T p(a), \T p(b), \F q(a), \T
  q(b), \F q(c), \F e_{\amp{\mathit{diff}[p, q]}}(a) \}$ suffices to
  correctly describe the input-output behavior. \qedhere
\end{example}

\leanparagraph{Functional Atoms}
When evaluating $\amp{g}[\vec{p}]$ with some functional $\amp{g}$ under
assignment~$\Assignment$, only one output tuple can be contained in
$\mathit{ext}(\amp{g}[\vec{p}], \Assignment)$,
formally: for all assignments~${\Assignment}$ and all $\vec{c}$, if
$\extsem{g}{\Assignment}{\vec{p}}{\vec{c}} = 1$ then $\extsem{g}{\Assignment}{\vec{p}}{\vec{c}'} = 0$
for all $\vec{c}' \not= \vec{c}$.
Therefore the following nogoods may be added right from the beginning.
\begin{definition}
\label{def:extlearn:functional}
The learning function for a functional external predicate $\amp{g}$ with input list
$\vec{p}$ in program $\Program$ under assignment $\Assignment$ is defined as
\begin{equation*}
\Lambda_f(\amp{g}[\vec{p}], \Assignment) =
		\left\{ \{ \T e_{\amp{g}[\vec{p}]}(\vec{c}),
		    \T e_{\amp{g}[\vec{p}]}(\vec{c'}) \}
			\mid \vec{c} \not= \vec{c'} \right\} \enspace .
\label{eqn:functional}
\end{equation*}
\end{definition}
However, our implementation of this learning rule does not generate all
pairs of output tuples beforehand.  Instead, it memorizes all generated
output tuples $\vec{c^i}$, $1 \leq i \leq k$ during evaluation of
external sources.  Whenever a new output tuple $\vec{c'}$ is added, it
also adds all nogoods which force previously derived output tuples
$\vec{c^i}$ to be false.
\begin{example}
	Consider the rules
	\begin{align*}
	  \mathit{out}(X) &\leftarrow \amp{\mathit{concat}}[A,x](X), \mathit{strings}(A), \mathit{dom}(X) \\[-1ex]
	  \mathit{strings}(X) &\leftarrow \mathit{dom}(X), \naf \mathit{out}(X)
	\end{align*}
	where $\ext{\mathit{concat}}{a,b}{c}$ is true iff string $c$ is
	the concatenation of strings $a$ and $b$,
	and observe that the external atom is involved in a cycle through negation.
	As the extension of the domain $\mathit{dom}$ can be large, many ground instances
	of the external atom are generated. The old evaluation algorithm guesses
        their truth values completely uninformed.
         E.g., $e_{\amp{\mathit{concat}}}(x,x,xx)$
	(the replacement atom of $\amp{\mathit{concat}}[A,x](X)$ with $A=x$ and $X=xx$,
	where $\mathit{dom}(x)$ and $\mathit{dom}(xx)$ are supposed to be facts)
	is in each guess set randomly to true or to false, independent of previous guesses.
	In contrast, with learning over external sources, the algorithm learns after
	the first evaluation that $e_{\amp{\mathit{concat}}}(x,x,xx)$ must be true.
        Knowing that $\amp{\mathit{concat}}$ is functional,
	all atoms $e_{\amp{\mathit{concat}}}(x,x,O)$ with $O \,{\neq}\, xx$
	must also be false. \qedhere
\end{example}

For the next result, let $\Program$ be a program which contains an
external atom of form $\amp{g}[\vec{p}](\cdot)$. %

\begin{lemma}
\label{lem:correctness:informedm}
For all assignments $\Assignment$,
\begin{inparaenum}[(i)]
\item the nogoods
$\Lambda_m(\amp{g}[\vec{p}], \Assignment)$
(Def.~\ref{def:extlearn:monotonic}) are correct wrt.~$\Program$, and
\item if $\amp{g}$ is functional, the nogoods
  $\Lambda_f(\amp{g}[\vec{p}], \Assignment)$
  (Def.~\ref{def:extlearn:functional}) are correct wrt.~$\Program$.
\end{inparaenum}
\end{lemma}
\extproof{
For monotonic external sources
we must show that negative input literals over monotonic parameters
can be removed from the learned
nogoods without affecting correctness.
For uninformed learning, we argued that
for output tuple $\vec{c} \in \mathit{ext}(\amp{g}[\vec{p}], \Assignment)$,
the replacement atom $e_{\amp{g}[\vec{p}]}(\vec{c})$ must not be be guessed false
if the input to $\amp{g}[\vec{p}](\vec{c})$ is $\Assignment|_{\vec{p}}$
under assignment $\Assignment$. However, as the output of $\amp{g}$ grows monotonically
with the extension of a monotonic parameter $p \in \vec{p_m}$,
the same applies for any $\Assignment'$ which is ``larger'' in $p$,
i.e., $\{ \T a \in \Assignment'|_{p} \} \supseteq \{ \T a \in \Assignment|_{p} \}$
and consequently $\{ \F a \in \Assignment'|_{p} \} \subseteq \{ \F a \in \Assignment|_{p} \}$.
Hence, the negative literals are not relevant
wrt. output tuple $\vec{c}$ and can be removed from the nogood.

For functional $\amp{g}$, we must show that the nogoods
$\left\{ \{ \T e_{\amp{g}[\vec{p}]}(\vec{c}),
    \T e_{\amp{g}[\vec{p}]}(\vec{c'}) \} \mid \vec{c} \not = \vec{c'} \right\}$
are correct.
Due to functionality, the
external source cannot return more than one output tuple
for the same input. Therefore no such guess
can be an answer set as it is not compatible.
Hence, the nogoods do not eliminate possible answer sets.
\qedhere
}

\leanparagraph{User-defined Learning}
In many cases the developer of an external atom has more information
about the internal behavior. This allows for defining more
effective nogoods. It is therefore beneficial to give the
user the possibility to customize learning functions.
Currently, user-defined functions
need to directly specify the learned nogoods.
The development of a user-friendly language for writing
learning functions is subject to future work.

\begin{example}
Consider the program
\begin{align*}
  r(X,Y) \vee \mathit{nr}(X,Y) &\leftarrow d(X), d(Y) \\
  r(V,W) &\leftarrow \ext{\mathit{tc}}{r}{V,W}, d(V), d(W)
\end{align*}
It guesses, for some set of nodes $d(X)$, all subgraphs of the
complete graph.
Suppose $\amp{\mathit{tc}}[r]$ checks if the edge selection $r(X,Y)$
is \underline{t}ransitively \underline{c}losed; if this is the case, the output is empty,
otherwise
the set of missing transitive edges is returned. For instance, if the
extension of $r$ is $\{ (a,b), (b,c) \}$, then the output
of $\amp{\mathit{tc}}$ will be $\{ (a,c) \}$, as this edge is missing
in order to make the graph transitively closed.
The second rule eliminates all
subgraphs which are not transitively closed.
Note that
$\amp{\mathit{tc}}$ is nonmonotonic.
The guessing program is
\begin{align*}
  r(X,Y) \vee \mathit{nr}(X,Y) &\leftarrow d(X), d(Y) \\
  r(V,W) &\leftarrow e_{\amp{\mathit{tc}}[r]}(V,W), d(V), d(W) \\
  e_{\amp{\mathit{tc}}[r]}(V,W) \vee \mathit{ne}_{\amp{\mathit{tc}}[r]}(V,W) &\leftarrow d(V), d(W)
\end{align*}
The naive implementation guesses for $n$ nodes all
$2^{\frac{n(n-1)}{2}}$ subgraphs and checks the transitive
closure for each of them, which is costly.
Consider the domain~$D = \{ a, b, c, d, e, f \}$. After checking one selection with $r(a,b), r(b,c), \mathit{nr}(a,c)$,
we know that \emph{no} selection containing these three literals will be transitively closed.
This can be formalized as a user-defined learning function.
Suppose we have just checked our first guess $r(a,b), r(b,c)$,
and $\mathit{nr}(x,y)$ for all other $(x,y) \in D \times D$.
Compared to the nogood learned by the general learning function, the nogood
$\{ \T r(a,b), \T r(b,c), \F r(a,c), \F e_{\amp{\mathit{tc}}[r]}(a,c) \}$
is a more general description of the conflict reason,
containing only relevant edges.
It is immediately violated and future guesses
containing $\{ \T r(a,b), \T r(b,c), \F r(a,c) \}$ are avoided. \qedhere
\end{example}

\begin{example}[Linearity]
\label{ex:smin}
A useful learning function for $\amp{\mathit{diff}}[p, q](X)$ is the following:
whenever an element is in $p$ but not in $q$, it belongs to the output
of the external atom.  This user-defined function works elementwise and
produces nogoods with three literals each. We call this property
\emph{linearity}.  In contrast, the naive learning function from the
Section~\ref{sec:learning:uninformed} includes the complete extensions
of $p$ and $q$ in the nogoods, which are less general. \qedhere
\end{example}

For user-defined learning, correctness of the learning function
must be asserted.

\section{Implementation and Evaluation}
\label{sec:impl}

We have integrated \clasp{} into our reasoner
\dlvhex{}; previous versions of \dlvhex{} used just \dlv{}.  In
order to learn nogoods from external sources we exploit \clasp{}'s SMT
interface, which was previously used for the special case of constraint
answer set solving and implemented in the \clingcon{}
system~\cite{geossc09a,os2012-tplp}.  We compare three configurations: \dlvhex{}
with \dlv{} backend, \dlvhex{} with (conflict-driven)~\clasp{} backend
but without EBL, and \dlvhex{} with \clasp{} backend and EBL.

For our experiments we used variants of the above examples, the \dlvhex{}
test
suite,
and default reasoning over
ontologies.
It appeared that learning has high potential to reduce the number of
candidate models.
Also the number of total variable assignments and backtracks during search decreased
drastically in many cases.
This
suggests that
candidate rejection often needs only
parts of interpretations
and
is possible early in the evaluation.
All benchmarks were carried out on a machine with two 12-core AMD
Opteron 6176 SE CPUs and 128 GB RAM, running Linux and using
\clasp{} 2.0.5 and \dlv{} Dec 21 2011 as solver backends.
For each benchmark instance, the average of three runs was calculated,
having a timeout of 300 seconds, and a memout of 2 GB for each run.  We
report runtime in seconds; gains and speedups are given as a factor.

\leanparagraph{Set Partitioning}
The following program partitions a set $S$ into two subsets
$S_1,S_2\subseteq S$
such that $\lvert S_1 \rvert \le 2$. The partitioning criterion is expressed
by two rules for $S_1 = S \setminus S_2$ and~$S_2 = S \setminus S_1$.
The implementation is by the use of external atom $\amp{\mathit{diff}}$
(cf. Example~\ref{ex:learningmonotonicity}):
\begin{align*}
	\mathit{dom}(c_1).\ &\dotsb \ \mathit{dom}(c_n). \\[-0.5ex]
	\mathit{nsel}(X) &\leftarrow \mathit{dom}(X), \ext{\mathit{diff}}{\mathit{dom}, \mathit{sel}}{X}. \\[-0.5ex]
	\mathit{sel}(X) &\leftarrow \mathit{dom}(X), \ext{\mathit{diff}}{\mathit{dom}, \mathit{nsel}}{X}. \\[-0.5ex]
	&\leftarrow \mathit{sel}(X), \mathit{sel}(Y), \mathit{sel}(Z), X \neq Y, X \neq Z, Y \neq Z.
\end{align*}
The results in Table~\ref{fig:setmin}
compare the run of the reasoner
with different configurations for computing
\begin{inparaenum}[(i)]
\item\label{eq:all} all models
resp.
\item\label{eq:first} the first model.
\end{inparaenum}
In both cases, using the conflict-driven \clasp{} reasoner instead of \dlv{}
as backend already improves efficiency.
Adding EBL leads to a further improvement:
in case~\eqref{eq:first}, the
formerly exponentially growing runtime becomes almost constant.
When computing all answer sets, the runtime
is still exponential
as exponentially many subset choices  must be considered
(due to the encoding);
however, also in this case many of them can be pruned early
by learning, which makes the runtime appear linear for the shown range of instance sizes.
Moreover, our experiments show that the delay between the models decreases
over time when EBL is used (not shown in the table), while
it is constant without EBL due to the generation of additional
nogoods.

\leanparagraph{Default Reasoning over Description Logic Ontologies}
We consider now a more realistic scenario using the
DL-plugin~\cite{eiks2009-amai} for \dlvhex{}, which integrates
description logics (DL) knowledge bases and nonmonotonic logic programs.  The
DL-Plugin allows to access an ontology using the description logic
reasoner {\small\texttt{RacerPro} 1.9.0}
({\small\url{http://www.racer-systems.com/}}).  For our first experiment, consider
the program (shown left) and the terminological part of a DL knowledge base on the right:
\begin{align*}
	\mathit{birds}(X) &\leftarrow \mathit{DL}[\mathit{Bird}](X).
             & \mathit{Flier} &\sqsubseteq \lnot \mathit{NonFlier} \\[-0.5ex]
	\mathit{flies}(X) &\leftarrow \mathit{birds}(X), \naf \mathit{neg\_flies}(X).
             & \mathit{Penguin} &\sqsubseteq  \mathit{Bird} \\[-0.5ex]
	\mathit{neg\_flies}(X) &\leftarrow \mathit{birds}(X),
        \mathit{DL}[\mathit{Flier} \uplus \mathit{flies}; \neg
        \mathit{Flier}](X).
             & \mathit{Penguin} &\sqsubseteq  \mathit{NonFlier}
\end{align*}
This encoding realizes the classic Tweety bird example using %
DL-atoms (which is an alternative syntax for external atoms in this
example and allows to express queries over description logics in a more
accessible way). The ontology states that $\mathit{Flier}$ is disjoint
with~$\mathit{NonFlier}$, and that penguins are birds and do not fly;
the rules express that birds fly by default, i.e., unless the contrary
is derived. The program amounts to the $\Omega$-transformation of
default logic over ontologies to dl-programs~\cite{dek2009}, where the
last rule ensures consistency of the guess with the DL ontology.
If the assertional part of the DL knowledge base contains
$\mathit{Penguin}(\mathit{tweety})$, then
$\mathit{flies}(\mathit{tweety})$ is inconsistent with the given
DL-program %
($\mathit{neg\_flies}(\mathit{tweety})$
is derived by monotonicity of DL atoms and~$\mathit{flies}(\mathit{tweety})$ loses its support).
\begin{table}[t]
\caption{Benchmark Results (runtime in seconds, timeout 300s)}
\subfloat[Set Partitioning]{\label{fig:setmin}%
 \begin{minipage}[t]{0.575\textwidth}
 \vspace{0cm} %
  \footnotesize
    \begin{tabular}[t]{@{}r@{~~~}r@{}r@{}r@{~~}r@{}r@{}r@{}}
    \toprule\toprule
    \multicolumn{2}{l}{\# elements} & \multicolumn{2}{l}{all models} & \multicolumn{3}{c}{first model} \\
        & \dlv{} & \clasp{} & \clasp{} & \dlv{} & \clasp{} & \clasp{} \\
        &        & w/o EBL & w EBL   &        & w/o EBL & w EBL \\
    \midrule
1	&	0.07	&	0.08	&	0.07	&	0.08	&	0.07	&	0.07 \\
5	&	0.20	&	0.16	&	0.10	&	0.08	&	0.08	&	0.07 \\
10	&	12.98	&	9.56	&	0.17	&	0.56	&	0.28	&	0.07 \\
11	&	38.51	&	21.73	&	0.19	&	0.93	&	0.63	&	0.08 \\
12	&	89.46	&	49.51	&	0.19	&	1.69	&	1.13	&	0.08 \\
13	&	218.49	&	111.37	&	0.20	&	3.53	&	2.31	&	0.10 \\
14	&	---	&	262.67	&	0.28	&	8.76	&	3.69	&	0.10 \\[-1.5mm]
$\vdots$~	&	---	&	---	&	$\vdots$~~~	&	$\vdots$~~~	&	$\vdots$~~~	&	$\vdots$~~~ \\
18	&	---	&	---	&	0.45	&	128.79	&	62.58	&	0.12 \\
19	&	---	&	---	&	0.42	&	---	&	95.39	&	0.10 \\
20	&	---	&	---	&	0.54	&	---	&	91.16	&	0.11 \\
\bottomrule\bottomrule
\end{tabular}
\end{minipage}
}
\hspace{5mm}
\subfloat[Bird-Penguin]{\label{fig:tweety}%
 \begin{minipage}[t]{0.31\textwidth}
 \vspace{0cm} %
  \footnotesize
    \begin{tabular}[t]{@{}r@{~~~}r@{}r@{}r@{}}
    \toprule\toprule
      \multicolumn{4}{l}{\# individuals}  \\
          & \dlv{} & \clasp{} & \clasp{} \\
          &        & w/o EBL & w EBL   \\
    \midrule
1 & 0.50 & 0.15 & 0.14 \\
5 & 1.90 & 1.98 & 0.59 \\
6 & 4.02 & 4.28 & 0.25 \\
7 & 8.32 & 7.95 & 0.60 \\
8 & 16.11 & 16.39 & 0.29 \\
9 & 33.29 & 34.35 & 0.35 \\
10 & 83.75 & 94.62 & 0.42 \\
11 & 229.20 & 230.75 & 4.45 \\
12 & --- & --- & 1.10 \\[-1.5mm]
$\vdots$~ & --- & --- & $\vdots$~~~ \\
20 & --- & --- & 2.70 \\
    \bottomrule\bottomrule
  \end{tabular}
\end{minipage}
}

\bigskip

\subfloat[Wine Ontology]{\label{tab:wine}%
 \begin{minipage}[t]{0.42\textwidth}
  \vspace{0cm} %
  \footnotesize
    \begin{tabular}[b]{@{}r@{}r@{}r@{}r@{}r@{}}
    \toprule\toprule
    Instance & \multicolumn{2}{c}{concept completion} &    \multicolumn{2}{c}{gain} \\
             & \clasp{}  & \clasp{} & max & avg \\
          &      w/o EBL & w EBL \\
    \midrule
wine\_0	&	25	&	31	&	33.02	&	6.93 \\
wine\_1	&	16	&	25	&	16.05	&	5.78 \\
wine\_2	&	14	&	22	&	11.82	&	4.27 \\
wine\_3	&	4	&	17	&	10.09	&	4.02 \\
wine\_4	&	4	&	17	&	6.83	&	2.87 \\
wine\_5	&	4	&	16	&	5.22	&	2.34 \\
wine\_6	&	4	&	13	&	2.83	&	1.52 \\
wine\_7	&	4	&	12	&	1.81	&	1.14 \\
wine\_8	&	4	&	4	&	1.88	&	1.08 \\
    \bottomrule\bottomrule
    \end{tabular}
   \end{minipage}
}
\hspace{5mm}
\subfloat[MCS]{\label{tab:mcs}%
 \begin{minipage}[t]{0.28\textwidth}
  \vspace{0cm} %
  \footnotesize
    \begin{tabular}[t]{@{}r@{~~~}r@{}r@{}r@{}}
    \toprule\toprule
\multicolumn{4}{l}{\hspace{-1mm}\# contexts}\\
	&	\dlv{}	&	\clasp{} & \clasp{}  \\
                &               &        w/o EBL & w EBL   \\
    \midrule
3	&	0.07	&	0.05	&	0.04 \\
4	&	1.04	&	0.68	&	0.14 \\
5	&	0.23	&	0.15	&	0.05 \\
6	&	2.63	&	1.44	&	0.12 \\
7	&	8.71	&	4.39	&	0.17 \\
    \bottomrule\bottomrule
    \end{tabular}
   \end{minipage}
}
\end{table}
Note that defaults cannot be encoded in standard (monotonic) description
logics, which is achieved here by the cyclic interaction of DL-rules and
the DL knowledge base.

As all individuals appear in the extension of the predicate
$\mathit{flier}$, all of them  are considered simultaneously.
This requires a guess on the ability to fly for each individual
and a subsequent check, leading to a combinatorial
explosion.
Intuitively, however, the property can be determined for each individual independently. Hence, a query
may be split into independent subqueries,
which is achieved by our learning function for \emph{linear sources}
in Example~\ref{ex:smin}. %
The learned nogoods are smaller and more candidate models are
eliminated.  Table~\ref{fig:tweety} shows the runtime for different
numbers of individuals and evaluation with and without EBL.  The runs
with EBL exhibit a significant speedup, as they exclude many model
candidates, whereas the performance of the \dlv{} and the \clasp{}
backend without EBL is almost identical (unlike in the first example);
here, most of the time is spent calling the description logic
reasoner and not for the evaluation of the logic program.

The findings carry over to large ontologies (DL knowledge bases) used in real-world
applications.  We did similar experiments with a scaled version of the
wine ontology ({\small\url{http://kaon2.semanticweb.org/download/test_ontologies.zip}}).
The instances differ in the size of the ABox (ranging from $247$
individuals in wine\_$0$ to $20007$ in wine\_${8}$) and in several
other parameters (e.g., on the number of concept inclusions and concept
equivalences;~\citeN{ms2006} describe the particular
instances~wine\_$i$).  We implemented a number of default rules using an
analogous encoding as above: e.g., wines not derivable to be dry are
not dry,
wines which are not sweet are assumed to be dry,
wines are white by default unless they are known to be red.
Here, we discuss the results of the latter scenario.
The experiments
classified the wines in the~$34$ main concepts of the ontology (the
immediate subconcepts of the concept $\mathit{Wine}$, e.g.,
$\mathit{DessertWine}$ and $\mathit{ItalianWine}$),
which have varying numbers of known concept memberships
(e.g., ranging from $0$ to $43$, and $8$ on average, in wine\_$0$)
and percentiles of red wines among them (from~$0\%$ to $100\%$, and
$47\%$ on average).
The results are summarized in Table~\ref{tab:wine}.
There, entries for concept completion state the number of classified concepts.
Again, there is
almost no difference between the~\dlv{} and the~\clasp{} backend without
EBL, but EBL leads to a significant improvement for most concepts and
ontology sizes. E.g., there is a gain for $16$ out of the $34$
concepts of the wine\_$0$ runs, as EBL can exploit linearity.
Furthermore, we observed that $6$ additional instances can be solved
within the $300$ seconds time limit. %
If a concept could be classified both with and without EBL, we could
observe a gain of up to $33.02$ (on average $6.93$).  As expected,
larger categories profit more from EBL as we can reuse learned nogoods
in these instances.

Besides $\Omega$, \citeN{dek2009} describe other transformations of
default rules over description logics. Experiments with this
transformations revealed that the structure of the resulting
\hex-programs prohibits an effective reuse of learned nogoods.  Hence,
the overall picture does not show a significant gain with EBL for
these encodings, we could however still observe a small improvement for
some runs.

\leanparagraph{Multi-Context Systems (MCS)}
MCS~\cite{be2007} is a formalism for interlinking multiple
knowledge-based systems (the contexts).  \citeN{efsw2010-kr} define
\emph{inconsistency explanations (IE)}\/ for MCS, and present a system
for finding such explanations on top of \dlvhex{}. In our benchmarks we
computed explanations for inconsistent multi-context systems with $3$ up
to $7$ contexts.  For each number we computed the average runtime over
several instances with different topologies (tree, zigzag, diamond), which
were randomly created with an available benchmark generator, and report
the results in Table~\ref{tab:mcs}.

Unlike in the previous benchmark we could already observe a speedup
of up to~$1.98$ when using \clasp{} instead of the \dlv{} backend.
This is because of two reasons: first,
\clasp{} is more efficient than \dlv{} for the given problem,
and second, \clasp{} was tightly integrated into \dlvhex{}, whereas using
\dlv{} requires interprocess communication. %
However, the most important aspect is again EBL, which leads to a further significant speedup
with a factor of up to $25.82$ compared to \clasp{} without EBL.

\leanparagraph{Logic Puzzles}
Another experiment concerns logic puzzles. We
encoded \emph{Sudoku} as a \hex{}-program, such that the logic
program makes a guess of assignments to the fields
and an external atom is used for verifying the
answer. In case of a negative verification result, the external atom indicates
by user-defined learning rules
the reason of the inconsistency, encoded a pair of assignments to fields
which contradict one of the uniqueness rules.

As expected, all instances times out without EBL, because the logic
program has no information about the rules of the puzzle and
blindly guesses all assignments, which are subsequently checked by the
external atom. But with EBL, the Sudoku instances could be solved in
several seconds.

More details on the experiments and links to benchmarks
and benchmark generators can be found at
\url{http://www.kr.tuwien.ac.at/research/systems/dlvhex/experiments.html}.

\section{Discussion and Conclusion}
\label{sec:conclusion}

The basic idea of our algorithm is related to constraint ASP solving
presented in~\citeN{geossc09a}, and~\citeN{os2012-tplp}, which is realized in the
\clingcon{} system. External atom evaluation in our algorithm can
superficially be regarded as constraint propagation.
However, while both,\citeN{geossc09a} and~\citeN{os2012-tplp}, consider a
particular application, we deal with a more abstract interface to
external sources.  An important difference between \clingcon{} and EBL
is that the constraint solver is seen as a black box, whereas we exploit
known properties of external sources.  Moreover, we support
\emph{user-defined learning}, i.e., customization of the default
construction of conflict clauses to incorporate knowledge about the
sources, as discussed in Section~\ref{sec:learning}.  Another difference
is the construction of conflict clauses.  ASP with CP has special
constraint atoms, which may be contradictory, e.g.,~$\T (X > 10)$ and
$\T (X = 5)$.  The learned clauses are sets of constraint literals,
which are kept as small as possible.  In our algorithm we have usually
\emph{no} conflicts between ground external atoms as output atoms are
mostly independent of each other (excepting e.g.\ functional
sources). Instead, we have a strong relationship between the input and
the output.  This is reflected by conflict clauses which usually consist
of (relevant) input atoms and the negation of one output atom. As in
constraint ASP solving,  the key for efficiency is keeping conflict
clauses small.

We have extended conflict-driven ASP solving techniques from ordinary
ASP %
to \hex-programs, which allow for using external atoms
to access external sources. Our approach uses two types of learning.
The classical type is conflict-driven clause learning, which derives
conflict nogoods from conflict
situations while the search tree is traversed. Adding such nogoods
prevents the algorithm from running into similar conflicts again.

Our main contribution is
a second type of learning which
we call \emph{external behavior learning} (EBL). Whenever external atoms are
evaluated, further nogoods may be added which capture parts of the
external source behavior.
In the simplest case these nogoods encode that a certain input to
the source leads to a certain output. This default learning function
can be customized to learn shorter or more general nogoods.
Customization is either done explicitly by the user, or learning functions
are derived automatically from known properties of external atoms,
which can be stated either on the level of external predicates or
on the level of atoms.
Currently we exploit monotonicity
and functionality.

Future work includes the identification of further properties which allow for
automatic derivation of learning functions.
We further plan the development of a user-friendly language for writing
user-defined learning functions. Currently,
they require to specify the learned nogoods by hand. It may be
more convenient to write
rules that
a certain input to an external source leads to a certain output,
in (a restricted variant of) ASP or a more convenient language. The
challenge is that evaluation of learning rules introduces additional
overhead, hence there is another tradeoff between costs and benefit of
EBL.  Finally, also the development of heuristics for lazy evaluation
of external sources is subject to future work.

\ifinlineref

\else
\bibliographystyle{acmtrans}
\bibliography{hexcdcl-bib}
\fi

\ifextended
\input{appendix}
\fi

\end{document}